# A Strong Feature Representation for Siamese Network Tracker

Zhipeng Zhou[a,b], Rui Zhang[a,b], Dong Yin[a,b]

[a]School of Information Science and Technology, USTC, Hefei, Anhui 230027, China
[b]Key Laboratory of Electromagnetic Space Information of CAS, Hefei, Anhui 230027, China

## Abstract

Because AlexNet is too shallow to form a strong feature representation, the tracker based on the Siamese network have an accuracy gap compared with state-of-the-art algorithms. To solve this problem, this paper proposes a tracker called SiamPF. Firstly, the modified pre-trained VGG16 network is fine-tuned as the backbone. Secondly, an AlexNet-like branch is added after the third convolutional layer and merged with the response map of the backbone network to form a preliminary strong feature representation. And then, a channel attention block is designed to adaptively select the contribution features. Finally, the APCE is modified to process the response map to reduce interference and focus the tracker on the target. Our SiamPF only used ILSVRC2015-VID for training, but it achieved excellent performance on OTB-2013 / OTB-2015 / VOT2015 / VOT2017, while maintaining the real-time performance of 41FPS on the GTX 1080Ti.

## 1. Introduction

Visual tracking is one of the most fundamental topics in computer vision. It has great demands in many public occasions such as surveillance system, self-driving cars, etc.

Many tracking methods were studied in recent years. They were mainly based on correlation filter framework or deep learning framework. Correlation filter was introduced to computer vision by David S. Bolme [1] who proposed its use for object tracking with MOSSE. Henriques J F proposed a method called CSK [2], which developed the intensive sampling and the kernel trick in the MOSSE. Furthermore, he exploited multi-channel HOG feature into KCF [3], which was an enhanced vision of CSK. Danelljan M [4] developed CSK with multi-channel color names(CN) feature. Due to their good performances, HOG and CN were becoming the most popular hand-craft features in recent years. Combining with features extracted from CNN, the correlation filter based methods such as DeepSRDCF [5], C-COT [6], ECO [7] were more robust while their complexity was greatly increased that could hardly meet the real-time property. An end-to-end tracking network was carried out as the development of deep learning. GOTURN[8] learned a generic relationship between object motion and appearance. SANet[9] used self-structure information of object to distinguish it from distractors. It utilized RNN to model object structure, and incorporate it into CNN to improve its robustness to similar distractors. TCNN[10] employed CNNs to represent target appearances, where multiple CNNs collaborated to estimate target states and determined the desirable paths for online model updates in the tree. These methods are either time-consuming or unsatisfactory in performance. Siamese network based trackers strike a balance between performance and speed. For example, SiamFC [11] used a fully convolutional Siamese network to get the score of template image beyond search image. It operated at frame-rates beyond real-time, despite its extreme simplicity, achieved state-of-the-art performance.

In this paper, inspired by transfer learning, we utilize the pretrained model (VGG16) trained on ImageNet as backbone, and fine-tuning it in different stage to get deep semantic information and multi-layer features. Furthermore, we design a channel attention block to force Siamese network to focus on object similarity metric. At last, like what has been done in CF trackers, we adopt modified APCE called APCEP to suppress the distractions. By testing on OTB2013 /OTB2015/VOT2015/VOT2017, our tracker achieve state-of-the-art performance while remains a real-time performance of 41FPS on the GTX 1080Ti. Some visual comparisons are shown in Figure 1.

The rest of the paper is organized as follows. Related work is introduced in Section 2. The proposed architecture and corresponding configurations are introduced in Section 3. While Section 4 presents the experimental results on four datasets and Section 5 concludes this paper.

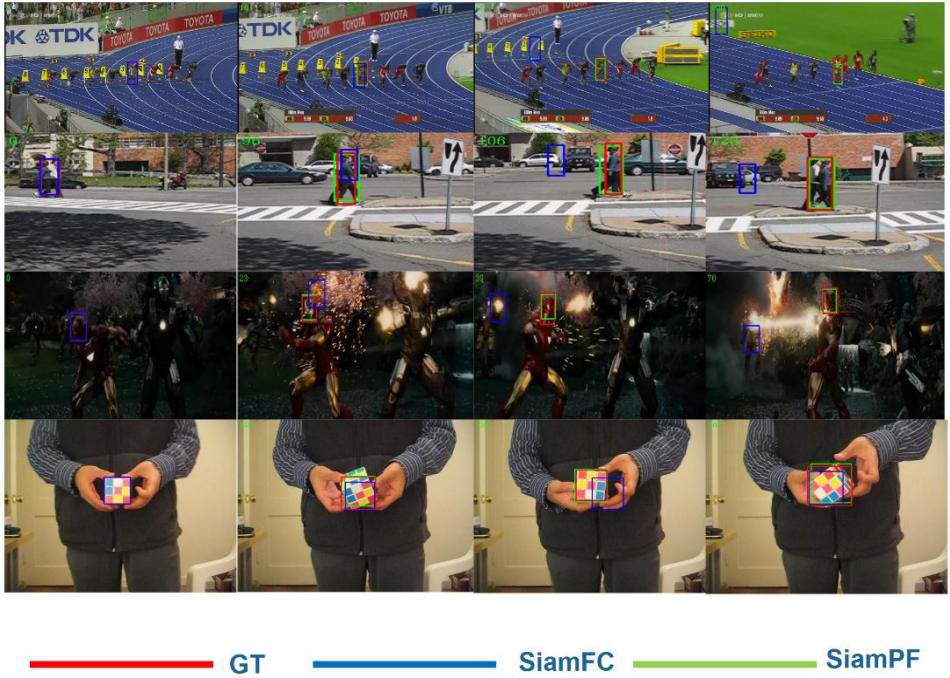

Figure 1. Comparing tracking results of SiamFC and our tracker SiamPF.

## 2. Related Work

### 2.1 Siamese Network Trackers

Siamese network, which has two branches for template and instance, is proposed to solve similarity matching problem. Some opinions regard visual tracking as a one-shot similarity metric task, since Siamese architectures are appropriate to address deep similarity learning problem, so Siamese network is extremely suitable for visual tracking. SiamFC[11] is the first approach to put the Siamese network integrating into visual tracking. After that, CFNet[12],DSiam[13] and SA-Siam[14] develop it by adding some operations on two branches. Borrowing from Region Proposal Network(RPN) in object detection, SiamRPN[15] adds two extra branches to do classification and regression. Recently, some works such as SiamVGG[16] focus on taking advantage of the capability of deeper neural network.

### 2.2 Average Peak-to Correlation Energy (APCE)

In Correlation Filter trackers, updating only happens with high confidence coefficient to prevent model from being polluted. To evaluate the confidence of tracking object and reflect the volatility of response map, APCE [19] was proposed. Nowadays, it is adopted in some CF trackers as a post-process on the response map to detect tracking failure. Its formulation can be concluded as Eq 1.

$$\text{APCE} = \frac{|F_{max} - F_{min}|^2}{mean(\sum_{w,h}(F_{w,h} - F_{min})^2)} \quad (1)$$

$F_{max}$ is the peak value, $F_{min}$ is the minimum value of response map, and $F_{w,h}$ is the value in the location $(w, h)$.

## 3. Proposed Method

The main idea of this paper is to build a strong feature extraction model without much training data. Figure 2 shows the major network proposed by this paper. AlexNet-like branch is added after the third layer of the modified VGG16. During training, the layers in modified VGG16 is frozen except last two layers. We also design an attention block to process the outputs of template in AlexNet-like branch.

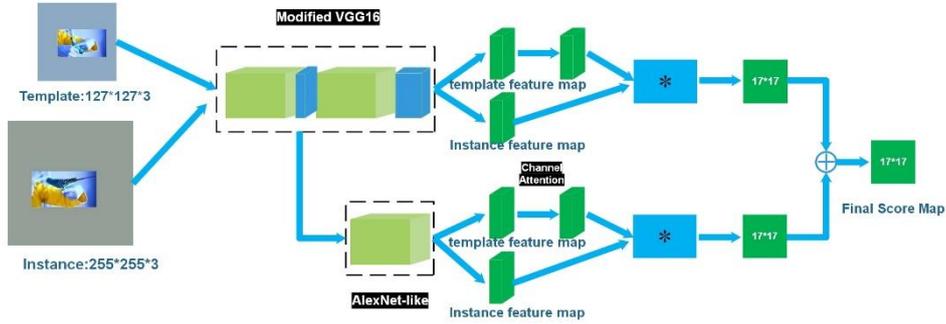

Figure 2. The network of our tracker. The instance and template images are fed into network as inputs. Extracting the features by a modified VGG16 model and an AlexNet-like branch which is introduced after conv3 of modified VGG16. The feature maps of template and instance are used for cross-correlation. The 17*17 score maps from both branches are combined to form the last score map.

### 3.1 Analysis of SiamFC

SiamFC employs a special way to acquire the annotation. The elements $y[u]$ of the final response map (corresponding to the final score map in Figure 2) are considered to belong to a positive example if they are within radius R of the centre $c$ (accounting for the stride k of the network) shown as Eq 2. In such way, every object gets the same ground-truth box though they have different size.

$$y[u] = \begin{cases} +1, & if\ k\|u - c\| \leq R \\ -1, & otherwise\ . \end{cases} \quad (2)$$

Examples projected from annotated response map are shown in Figure 3. From it, we can see SiamFC does not care about the edge information, it just focus on predicting the centre point of object and makes the extracted features less representative, which degrades the feature extraction of the whole network.

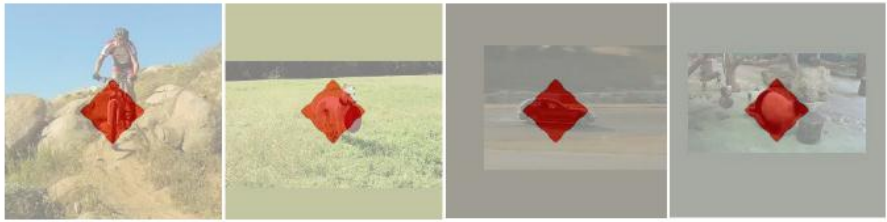

Figure 4. Labels projected from response map

### 3.2 Strong Feature Representation Backbone Network

As analysis in 3.1, the present operations in SiamFC do not fully take advantage of current training data and are hard to meet great performance. In order to obtain a strong feature representation, we choose pretrained model as our backbone network.

Main deep learning based trackers use modified AlexNet as their backbone. As a matching task, AlexNet is not deep enough to obtain high semantic. While in some deep networks like ResNet, some operations like padding would lead to object location preference. Therefore, we utilize modified VGG16 network pretrained on ImageNet and fine-tuning it in different stage to obtain multi-layer features including semantic and appearance information. We fine-tune modified VGG16 on last two layers and add another AlexNet-like branch after third layer of modified VGG16. The structure of modified VGG16 and AlexNet-like branch is detailed in Table 1.

Table 1. Structures of modified VGG16 and AlexNet-like branch. All the convolutional layers are integrated with BatchNorm and ReLU except the last one working for generating outputs. 'K-size' means 'Kernel Size', 'C-in' means 'Channel in', 'C-out' means 'Channel out'.

| modified VGG16 | | | | | AlexNet-like branch | | | | |
|---|---|---|---|---|---|---|---|---|---|
| Type | K-size | C-in | C-out | Stride | Type | K-size | C-in | C-out | Stride |
| Conv1 | 3 | 3 | 64 | 1 | Conv1 | 5 | 128 | 256 | 1 |
| Conv2 | 3 | 64 | 64 | 1 | MaxPool1 | 3 | | | 2 |
| MaxPool1 | 2 | | | 2 | Conv2 | 3 | 256 | 384 | 1 |
| Conv3 | 3 | 64 | 128 | 1 | Conv3 | 3 | 384 | 256 | 1 |
| Conv4 | 3 | 128 | 128 | 1 | Conv4 | 3 | 256 | 256 | 1 |
| MaxPool2 | 2 | | | 2 | | | | | |
| Conv5 | 3 | 128 | 256 | 1 | | | | | |
| Conv6 | 3 | 256 | 256 | 1 | | | | | |
| Conv7 | 3 | 256 | 256 | 1 | | | | | |
| MaxPool3 | 2 | | | 2 | | | | | |
| Conv8 | 3 | 256 | 512 | 1 | | | | | |
| Conv9 | 3 | 512 | 512 | 1 | | | | | |
| Conv10 | 3 | 512 | 512 | 1 | | | | | |
| Conv11 | 3 | 512 | 256 | 1 | | | | | |

We define the AlexNet-like branch and VGG16 branch as $\varphi_A(\cdot)$ and $\varphi_V(\cdot)$ respectively.

After generating the feature map through the fully convolutional Siamese network, we compute cross-correlation of two input images, which is define as Eq 3 and Eq 4, where $z$ and $x$ represent the exemplar image and instance image, respectively. $\varphi_*(\cdot)$ represents a convolutional embedding function and $b\mathbb{1}$ denotes the bias with value $b \in \mathbb{R}$.

$$f_A(x_T, x_I) = \varphi_A(z) * \varphi_A(x) + b\mathbb{1} \quad (3)$$
$$f_V(x_T, x_I) = \varphi_V(z) * \varphi_V(x) + b\mathbb{1} \quad (4)$$

And the final response map S is defined as Eq 5.

$$S = \lambda * f_V(x_T, x_I) + (1-\lambda) f_A(x_T, x_I) \quad (5)$$

$\lambda$ is a hyper-parameter to balance two response map.

### 3.3 Attention Strengthened

As a matching task, features of exemplar image play an important role. There exists many background noises which would mislead tracker to drift in the exemplar image. To further strengthen the current object feature representation, we add an attention block to the output feature map of exemplar images. We have different process in two branches. We adopt a channel attention block (see Figure 4) for AlexNet-like branch while takes no measures for VGG16 branch. From our experiment, adding an attention block in VGG16 branch would hurt the performance, mainly because the VGG16 output feature map has high semantic information, attention block would hurt the relationship among these channels.

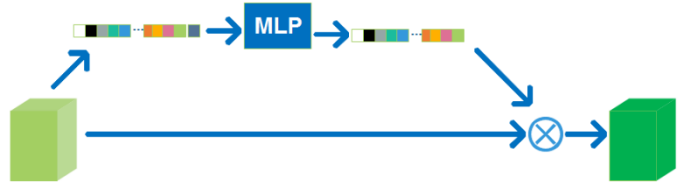

Figure 4. Channel Attention Block.

### 3.4 Post-process

Refer to the correlation filter based trackers, our tracker also benefits from post-process. Since the resized feature map has the same resolution as the input, APCE would be effective to depress distractors. But APCE is still too compact to distinguish success and failure during tracking. So we modify Eq 1 to enlarge the output range of APCE. It can be concluded in Eq 6.

$$\text{APCEP} = \left( \frac{F_{max}^2 - F_{min}^2}{mean\left(\sum_{w,h}(F_{w,h}-F_{min})^2\right)} \right)^2 \qquad (6)$$

Figure 5 shows the comparison between APEC and APCEP in normal and occluded situation.

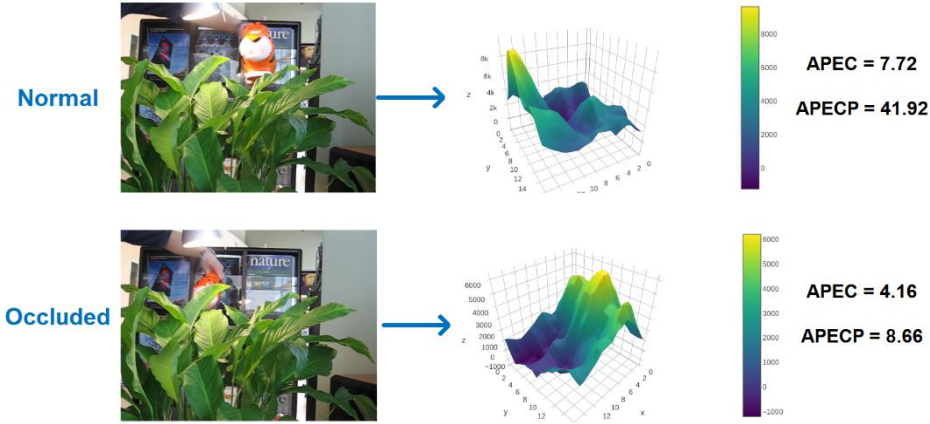

Figure 5. Comparison of APCE and APCEP

From Figure 5, we can see in the normal scene, APCEP gets a much higher score, but in occluded scene, APCEP degrade a lot and approximates as APCE.

## 4. Experiment

Experiments are performed on four public datasets: OTB2013, OTB2015, VOT2015, VOT2017. OTB contains 100 videos while VOT contains 60 videos.

### 4.1 Implementation Details

The experimental platform in this paper is CPU: i7-7700, GPU: GTX1080Ti, Memory: 16G, Operating system: Ubuntu 16.04.We train our model with straightforward SGD using PyTorch. Training is performed with 50 epochs. The gradients for each iteration are estimated using mini-batches of size 8, and the learning rate is annealed at every 20 epochs from $10^{-1}$ to $10^{-3}$. $\lambda$ is set to 0.75. The other hyper-parameters are set the same as SiamFC. The code would be released at https://github.com/zzpustc/SiamPF soon.

### 4.2 Evaluation

#### 4.2.1 Results on OTB2013/OTB2015

OTB[20] contains 100 sequences that are collected from commonly used tracking sequences. The evaluation is based on two metrics: success and precision plot. When the overlap between groundtruth box and generated box is larger than a given threshold in a certain frame, we call it a successful frame. The precision plot shows the percentages of frames that tracking results are within 20 pixels from the target. So we test our tracker on the benchmark comparing with MCPF[21], ECOhc[7], CFNet[12], SiamFC[11], Staple[22]. Figure 6 and Figure 7 show the results on OTB2013, OTB2015 respectively.

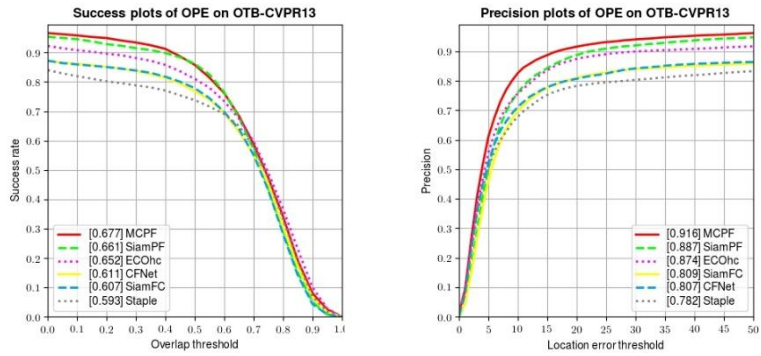

Figure 6. Success plot and precision plot of SiamPF on the OTB2013

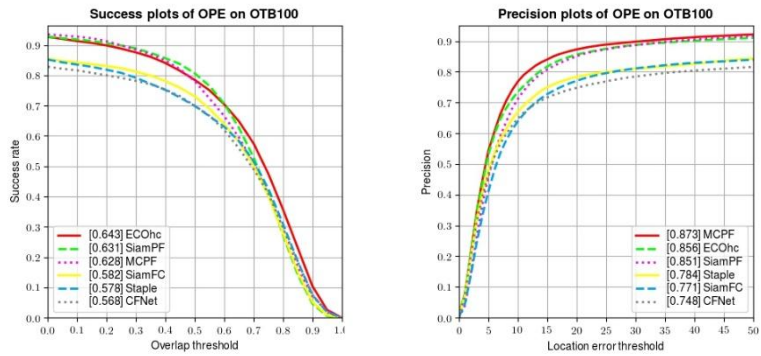

Figure 7. Success plot and precision plot of SiamPF on the OTB2015

From Figure 6 and Figure 7, we can see that SiamPF rank top among these trackers both in success plot and precision plot.

**4.2.2 Results on VOT2015/VOT2017**

VOT[23] contains totally 60 sequences. Its metrics consist of accuracy and robustness. And the overall performance is evaluated using Expected Average Overlap(EAO) which takes account of both accuracy and robustness. Besides, a new real-time experiment is conducted. Figure 8, Figure 9 show order of trackers (MDNet[24], DeepSRDCF[25], EBT[26], srdcf[27], sPST[28], scebt[29], nsamf[30], struck[31],, CFCF[32], mcct[33], csr[34], MCPF[21], CRT[35], ECOhc[7] etc.) by evaluating their EAO on VOT2015, VOT2017. Table 2, Table 3 contain more qualitative details (Red means ranking first, blue means ranking second, green means ranking third).

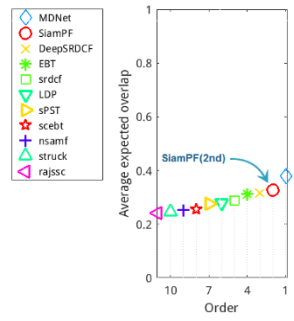

Figure 8. Expected overlap scores of SiamPF on the VOT2015, larger is better.

Table 2. Detail information about several state-of-the-art trackers performance on the VOT2015

| Tracker | EAO | Accuracy | Failures | A-R rank |
| --- | --- | --- | --- | --- |
| MDNet | 0.3783 | 0.5991 | 13.1519 | 1 |
| DeepSRDCF | 0.3181 | 0.5565 | 16.9525 | 3 |
| EBT | 0.3130 | 0.4596 | 15.3702 | 4 |
| srdcf | 0.2877 | 0.5529 | 21.2642 | 5 |
| LDP | 0.2785 | 0.4841 | 23.8973 | 6 |
| sPST | 0.2767 | 0.5479 | 26.2529 | 7 |
| scebt | 0.2548 | 0.5423 | 31.8157 | 8 |
| nsamf | 0.2536 | 0.5246 | 25.6161 | 9 |
| struck | 0.2458 | 0.4537 | 27.1530 | 10 |
| **SiamPF** | **0.3263** | **0.5905** | **18.6719** | **2** |

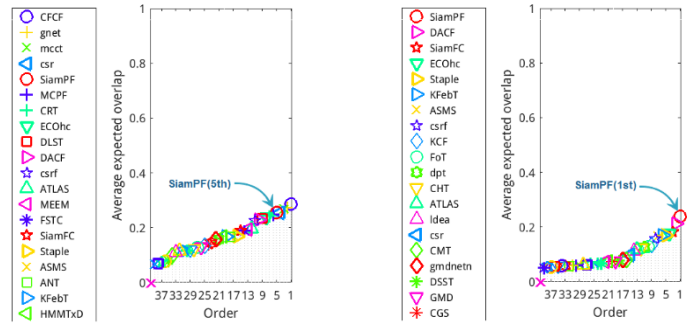

(a)          (b)

Figure 9. Expected overlap scores of SiamPF on the VOT2017, larger is better. (a) is the baseline experiment, (b) is the real-time experiment.

Table 3. Detail information about several state-of-the-art trackers performance on the VOT2017

| Tracker | EAO(baseline) | Accuracy | Failures | A-R rank | EAO(real-time) |
|---------|---------------|----------|----------|----------|----------------|
| CFCF | 0.2857 | 0.5049 | 19.6495 | 1 | 0.0587 |
| gnet | 0.2737 | 0.4999 | 17.3674 | 2 | 0.0599 |
| mcct | 0.2703 | 0.5228 | 19.4526 | 3 | 0.0605 |
| csr | 0.2561 | 0.4846 | 23.5731 | 4 | 0.0993 |
| MCPF | 0.2478 | 0.5035 | 25.9600 | 6 | 0.0602 |
| CRT | 0.2441 | 0.4639 | 21.0611 | 7 | 0.0683 |
| ECOhc | 0.2384 | 0.4893 | 28.7674 | 8 | 0.1767 |
| DLST | 0.2332 | 0.5038 | 24.6046 | 9 | 0.0568 |
| DACF | 0.2285 | 0.4494 | 25.2403 | 10 | 0.2120 |
| **SiamPF** | **0.2554** | **0.5006** | **23.3362** | **5** | **0.2376** |

From the present results above, we can see that different trackers have different advantages, but SiamPF is always among the top tier over all the evaluation metrics. Specially, SiamPF rank 1st in the real-time experiment on VOT2017, which means our tracker has the best balance on EAO and speed.

**4.3 Ablation Analysis**

In this section, we analyse the influence of each operation to the final performance, and it can be concluded in Table 4. From Table 4, it can be observed that pretrained model bring the greatest promotion. AlexNet-like branch also carries out 0.8% increase. Channel attention block and APCEP perform differently in two datasets but still benefit for our tracker.

Table 4. Influence analysis of each operation. ✓ means adding this operation into tracker.

| Frozen Pretrained Model | AlexNet-like branch | Channel Attention | APCEP | OTB2013 AUC | OTB2015 AUC |
|---|---|---|---|---|---|
|  |  |  |  | 0.6130 | 0.5958 |
| ✓ |  |  |  | 0.6431 | 0.6148 |
| ✓ | ✓ |  |  | 0.6511 | 0.6274 |
| ✓ | ✓ | ✓ |  | 0.6565 | 0.6348 |
| ✓ | ✓ | ✓ | ✓ | 0.6610 | 0.6310 |

# 5. Conclusion

In this work, we propose a tracker called SiamPF, which fintunes pretrained VGG16 model in different stage to obtain multi-layer features. We design an attention block to strengthen the feature representation. At last, we modified APCE to process the score map. In this way, SiamPF achieve top performance on OTB2013/OTB2015/VOT2015/VOT2017. But there still exists many problems remains to be solved. The shallow features are not fully explored. Spatial-temporal property should be considered in long-term tracking. Our next work would focus on these problems.